\documentclass{article}
\usepackage{spconf,amsmath,graphicx}
\usepackage{flushend}

\title{FACIAL EMOTION RECOGNITION USING DEEP LEARNING}
\name{Ching-Da Wu and Li-Heng Chen}
\address{The University of Texas at Austin}
\begin{document}
%
\maketitle
\begin{abstract}
In this project, we aim to construct a system that captures real-world facial images through the front camera on a laptop. The system is capable of processing/recognizing the captured image and predict a result in real-time. In this system, we exploit the power of deep learning technique to learn a facial emotion recognition (FER) model based on a set of labeled facial images. Finally, experiments are conducted to evaluate our model using largely used public database.

\end{abstract}
\section{Introduction}
\label{sec:intro}

With the growing in the field of multimedia applications including social robots, medical treatment (detection of mental disorders), and video chat, recognizing facial expression has become an important but hard problem. As one of the most natural signal that human use to convay their emotion or intention, facial expression has been actively studied for a long time. In fact, most of the machine learning based FER algorithms comprise feature extraction (FE) and classification: Several popular FE techniques such as Gabor filtering\cite{Li2013,Liu2014}, Local Binary Pattern (LBP)\cite{Zhang14}, Principal Component Analysis (PCA)\cite{AP91} are used as the input of a classifier. Then, trained machine learning classifier such as Statistical Classification, Support Vector Machine (SVM)\cite{tu2018content} are employed to predict a result based on the feature vector extracted from a facial image.

\section{System Overview}
\label{sec:format}
As an introduction to this FER system, we briefly separate the whole system in three major functional blocks:
\subsection{Webcam Capturing}
The input of this system is defined images captured from camera. To access the front camera embedded in the laptop and process it, we utilize the VideoCapture library in OpenCV \cite{opencv_library} (Open Source Computer Vision Library: http://opencv.org), which allows us to capture image/video from cameras and use the information in real time. Additionally, input pre-processing is made before being used to predict facial emotion. This will be introduced later in section \ref{sec:OPT}.
\subsection{Prediction Model}
\begin{figure}[t]
  \centering
  \centerline{\includegraphics[width=8.5cm]{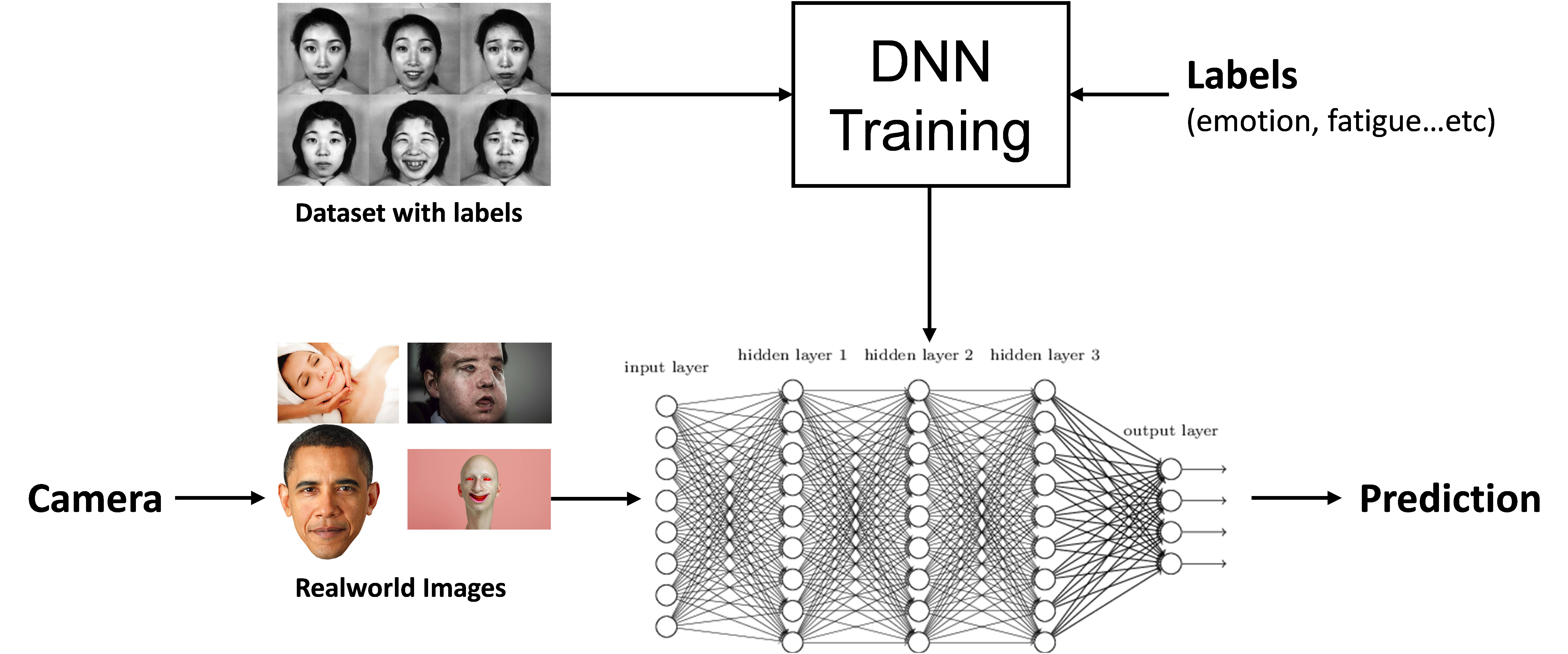}}
\caption{The framework of this system: a deep neural network (DNN) is trained using available image dataset with facial emotion label. The (pre-processed) input image is then fed into the pre-trained model, and its facial expression is predicted and displayed on the user interface.}
\label{fig:overview}
\end{figure}

As demonstrated in Fig. \ref{fig:overview}, the core function is a pre-trained Deep Neural Network model. Facial images with emotion label are collected to determine parameters of a pre-defined network architecture, and the trained model is evaluated with a plurality of test image which is a disjoint set to the training images. A more detailed introduction can be found in section \ref{sec:DL}.
\subsection{User Interface (UI) Design}
Displaying the result is also important. To demonstrate our system, we design a simple user interface that can render the recognition result of facial emotion on the input image and display it on the screen.

\begin{figure*}[htb]
  \centering
  \centerline{\includegraphics[width=16.5cm]{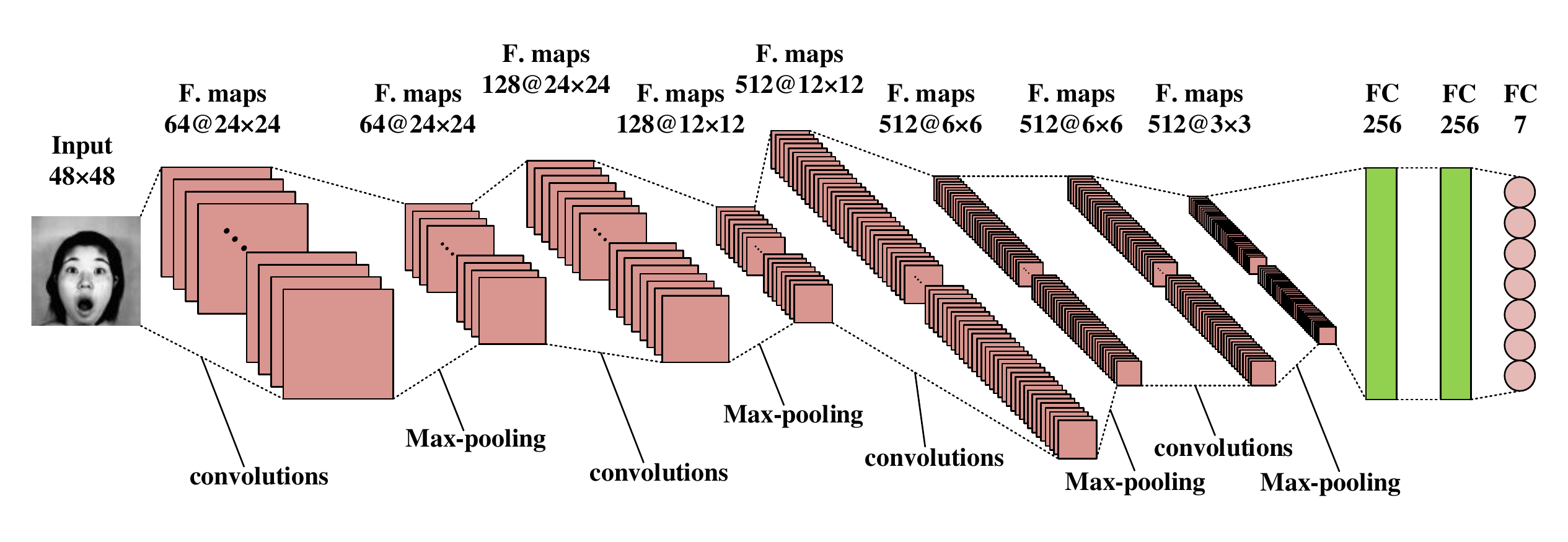}}
\caption{Illustration of our 4-layer built-from-scratch DNN model. Each layer is composed of convolutions, max-pooling, and a Rectified Linear Unit (ReLU) activation function. The parameterization of the convolutional layer is denoted as "Channels @ width $\times$ height".}
\label{fig:cnn4layer}
\end{figure*}

\section{Dataset Selection}
\label{sec:pagestyle}

In order to amass sufficient data for model training, we searched a plurality of public dataset on the internet as listed in Table \ref{table:data}. However, some of them may not be suitable for training a promising recognition model: For instance, JAFFE dataset \cite{jaffe97} collected data from only ten subjects, and all of them are female; FEI \cite{feiface} dataset only has two labels, Neutral and Smile, which may not give us a decent result as well. Considering the comprehension of dataset and our computational resource, we decide to use Kaggle dataset \cite{kaggle13} as our training data.

\begin{table}
  \centering

    \begin{tabular}{ | l | l | l | l | l |}
    \hline
    Name         & Class & Img    & Sub    & Format        \\ \hline
    JAFFE        & 7     & 213    & 10 (F) & 256x256 gray  \\
    Kaggle       & 7     & 28709  & N/A    & 48x48 gray    \\
    Yale Face    & 11    & 165    & 15     & 320x243 gray  \\
    CMU PIE      & 6     & 750000 & 28     & 3072x2048 rgb \\
    RAVDESS      & 8     & 7356   & 24     & 1280x720      \\
    FEI          & 2     & 2800   & 200    & 640x480       \\
    \hline
    \end{tabular}

  \caption{A list of public facial emotion recognition dataset: Name of dataset (Name), number of facial expression (Class), number of facial images (Img), number of subjects (Sub), and image format (Format) are presented in this table.}
  \label{table:data}
\end{table}

\section{Our Method: Deep Learning Treatment}
\label{sec:DL}

In this section, we introduce our deep learning model including the architecture, training process, and fine-tuning work:
\subsection{Architectural Choices}
\label{ssec:dlarch}
Convolutional Neural Networks (CNN) have been used extensively in the recent years for computer vision and neural language processing\cite{liu2019cyclicgen}. We have adopted many state-of-the-art CNN as the building blocks of our model architecture, such as VGG \cite{vgg}, ResNet \cite{res}, DenseNet \cite{dense} and MobileNet \cite{mobile}. Despite their decent prediction accuracy, it still takes too much execution time for some of these models on the testing phase. Therefore, we try to simplify the models and try to build up our own CNN model from scratch. In the project, we use a CNN with four convolutional blocks as described in Fig. 2. Each convolutional block is consist of a batch normalization, a convolutional layer with (N$\times$N) kernel size with a rectifier-linear-unit (ReLU) activation function, and a max pooling layer with (2$\times$2) size. The batch normalization is utilized to ensure the input data have zero mean and unit variance. All the convolution layers have (1$\times$1) stride. The four convolution layers have $\{64, 128, 512, 512\}$ kernels, respectively. Following the four convolution blocks are three fully-connected layers with sigmoid activation and output sizes of $\{256,256,7\}$. Dropout layers are placed before each of the three fully-connected layers with dropping probabilities $0.3$.

\subsection{Training Data Preprocessing}
\label{ssec:trainpreproc}
\begin{figure}[htb]
\begin{minipage}[b]{0.2\linewidth}
  \centering
  \centerline{\includegraphics[width=1.5cm]{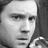}}
  \centerline{(a)}\medskip
\end{minipage}
\hfill
\begin{minipage}[b]{0.2\linewidth}
  \centering
  \centerline{\includegraphics[width=1.5cm]{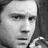}}
  \centerline{(b)}\medskip
\end{minipage}
\hfill
\begin{minipage}[b]{0.2\linewidth}
  \centering
  \centerline{\includegraphics[width=1.5cm]{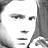}}
  \centerline{(c)}\medskip
\end{minipage}
\hfill
\begin{minipage}[b]{0.2\linewidth}
  \centering
  \centerline{\includegraphics[width=1.5cm]{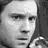}}
  \centerline{(d)}\medskip
\end{minipage}
\begin{minipage}[b]{0.2\linewidth}
  \centering
  \centerline{\includegraphics[width=1.5cm]{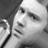}}
  \centerline{(e)}\medskip
\end{minipage}
\hfill
\begin{minipage}[b]{0.2\linewidth}
  \centering
  \centerline{\includegraphics[width=1.5cm]{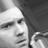}}
  \centerline{(f)}\medskip
\end{minipage}
\hfill
\begin{minipage}[b]{0.2\linewidth}
  \centering
  \centerline{\includegraphics[width=1.5cm]{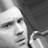}}
  \centerline{(g)}\medskip
\end{minipage}
\hfill
\begin{minipage}[b]{0.2\linewidth}
  \centering
  \centerline{\includegraphics[width=1.5cm]{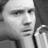}}
  \centerline{(h)}\medskip
\end{minipage}
\caption{Example of pre-processing a training patch in Kaggle2013 database: (a) Original patch (b)-(d) random brightness adjustment (e)-(h) random flipping/rotation}
\label{fig:traindataarg}
\end{figure}

Not only do we aim to train a prediction model with high accuracy, robustness of a model is also an important aspect. In real-world image capturing, rotation and scaling of a facial image occurs often. Moreover, brightness of an image varies rapidly according to the lighting condition. Unfortunately, most of the available databases do not take these factors into account.\\
To make an improvement on this, we randomly impose a plurality of pre-processing on each training patch. As shown in Fig. \ref{fig:traindataarg}, different level of image rotation and brightness adjustment are performed on an original image patch. With this treatment, the diversity as well as the amount of a dataset is significantly increased.

\subsection{Fine-Tuning of DNN Model}
\label{ssec:dltune}
To further improve the performance, we adopt a two-stage training procedure to fine-tune our DNN model. The first stage in fact has no difference to a normal training method. However, in the second stage, we fixed the parameters of the first four convolutional layers to the result we achieved in the first stage and re-train the last two fully-connected layers. Also, the learning rate for the second stage is set to be 10 times smaller than the first stage.

\section{Optimization in system level}
\label{sec:OPT}
Aside from training a DNN model with high enough prediction accuracy, we also deployed several algorithms to optimize the whole system and improve user experience. In fact, the input of this system (camera-captured images) does not have proper emotion label. Therefore, they cannot be jointly trained with the labeled dataset. Given this, the followings are empirically designed to fine-tune the whole system:  

\subsection{Input Denoising}
\label{ssec:denoise}

\begin{figure}[t]

\begin{minipage}[b]{.45\linewidth}
  \centering
  \centerline{\includegraphics[width=4.0cm]{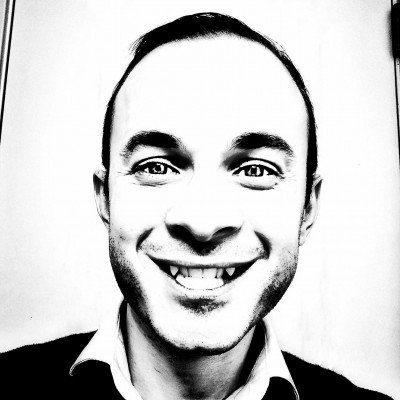}}
  \centerline{(a) An image in public dataset}\medskip
\end{minipage}
\hfill
\begin{minipage}[b]{0.45\linewidth}
  \centering
  \centerline{\includegraphics[width=4.0cm]{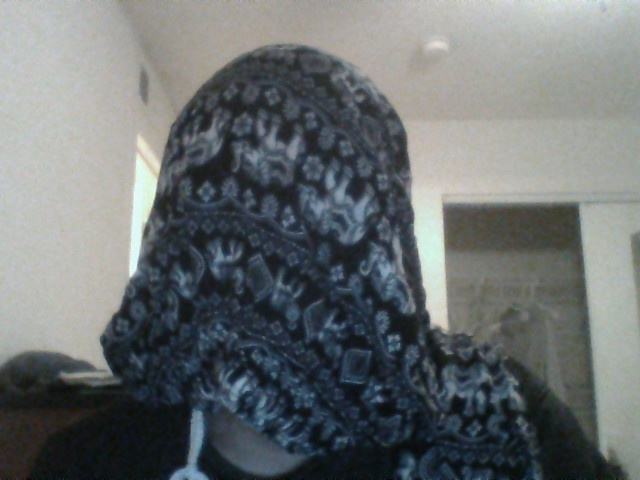}}
  \centerline{(b) Camera-captured image}\medskip
\end{minipage}
\caption{Example of images from different source}
\label{fig:noise}
\end{figure}

It is easy to observe that the images in the selected dataset have different property from the camera-captured images. One is that the image is corrupted by ISO noise after being captured by the front camera whereas the images in dataset are relatively pristine as shown in Fig \ref{fig:noise}.

To fill this gap, we embedded a de-noise module to mitigate the noise. By experiment, we believe a simple 3x3 Gaussian filter provides the best result for our system. Although there exists more powerful but sophisticated de-noise algorithms such as Non-Local Mean filter, integrating these computationally consuming algorithms into our pipeline may harm the frame rate of our system. In other word, the system cannot be executed in real time with these algorithms.

\subsection{Scene-Change Detection}
\label{ssec:scd}

\begin{figure}[t]
  \centering
  \centerline{\includegraphics[width=8.5cm]{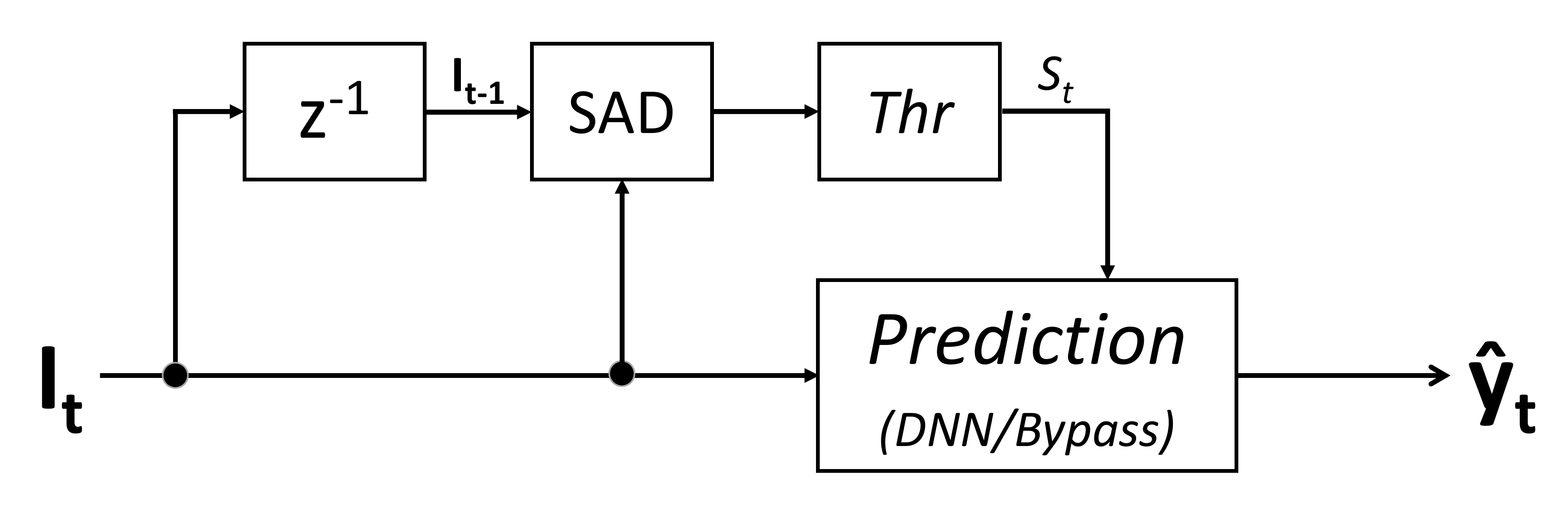}}
\caption{A simple scene change detection algorithm is used to filter out subtle changes between consecutive frames. This flowchart shows that SAD between $\mathbf{I_t}$ and $\mathbf{I_{t-1}}$ is calculated and being used to control the prediction. Notice that $z^{-1}$ means delaying by 1-frame. }
\label{fig:scenechange}
\end{figure}

Before feeding a processed image into the pre-trained DNN model, we designed a low-complexity scene-change detector. The DNN model is activated only when the current frame is detected as non-static scene. As shown in Fig. \ref{fig:scenechange}, the scene-change detector calculates sum of absolute difference (SAD) between the processed current frame $\mathbf{I_t}$ and the previous frame $\mathbf{I_{t-1}}$, where
\begin{align}
SAD(\mathbf{I_t}, \mathbf{I_{t-1}})= \|\mathbf{I_t}-\mathbf{I_{t-1}}\|_{1,1}
\nonumber \\ =\sum_{i,j}|I_t(i,j)-I_{t-1}(i,j)|
\end{align}
and the scene type $S_t$ is then determined by
\begin{equation}
S_t = \begin{cases}
1 \text{ (scene-change)} &\text{, if $SAD(\mathbf{I_t}, \mathbf{I_{t-1}}) > Thr$}\\
0 \text{ (static scene)} &\text{, otherwise}
\end{cases}
\end{equation}
note that the DNN model is bypassed (prediction remains the same) when $S_t=0$. To be more specific, the prediction result can be written as
\begin{equation}
\hat{y}_t = \begin{cases}
DNN(\mathbf{I_t})  &\text{, if $S_t=1$} \text{ (scene-change)}\\
\hat{y}_{t-1}      &\text{, if $S_t=0$} \text{ (static scene)}
\end{cases}
\end{equation}
Without this module, the prediction result may change rapidly even when the subject is not moving. 

\section{Experimental Results}
\label{sec:illust}
In this section, we report a more detailed experimental result of our model:

\subsection{Experiment on Different Architectures}
\label{ssec:archexp}
As mentioned previously, we conducted experiments through several existing DNN architectures and one built-from-scratch architecture to understand which model is the best fit to our training data. For the existing architectures, we remove its original fully-connected layers on the top, and replace it with two 1024 sized fully-connected layers and a 7-class output dense layer. In our training procedure, we randomly split the 32298 samples in the Kaggle2013 database to the proportion of Train$:$Validate$:$Test$=8:1:1$, and the prediction accuracy is calculated from the testing data. As shown in Table. \ref{table:modelsel}, the number of parameters, prediction accuracy, and run time is reported in detail. We can see that VGG16 performs the best in terms of accuracy, with acceptable run time. Also, we noticed that a deeper model, such as DenseNet or ResNet, can easily over-fit the training samples especially when the amount of data is not large enough. \\

\begin{figure}[tb]
\begin{minipage}[b]{.49\linewidth}
  \centering
  \centerline{\includegraphics[width=4.7cm]{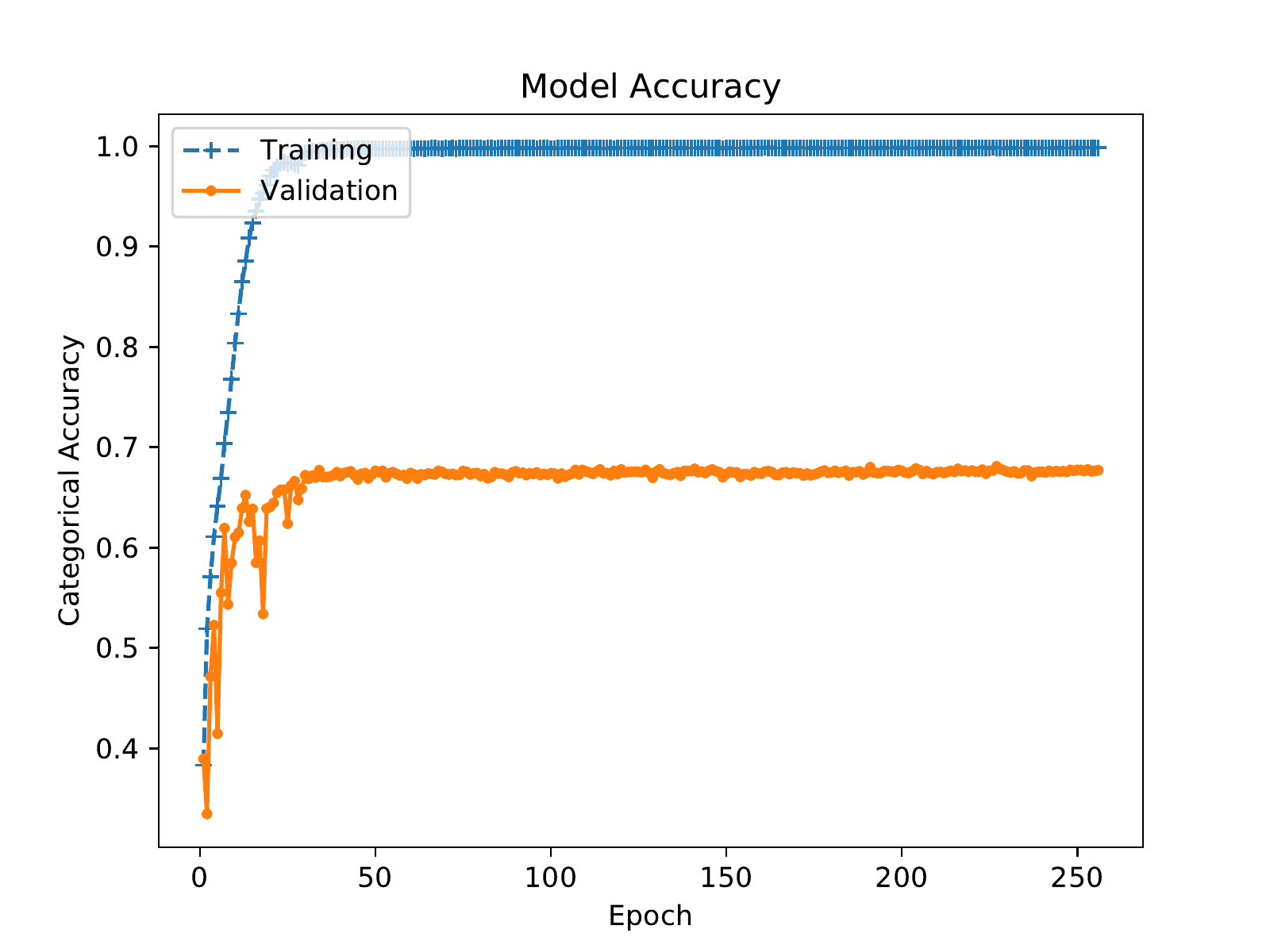}}
  \centerline{(a) VGG16}\medskip
\end{minipage}
\hfill
\begin{minipage}[b]{0.49\linewidth}
  \centering
  \centerline{\includegraphics[width=4.7cm]{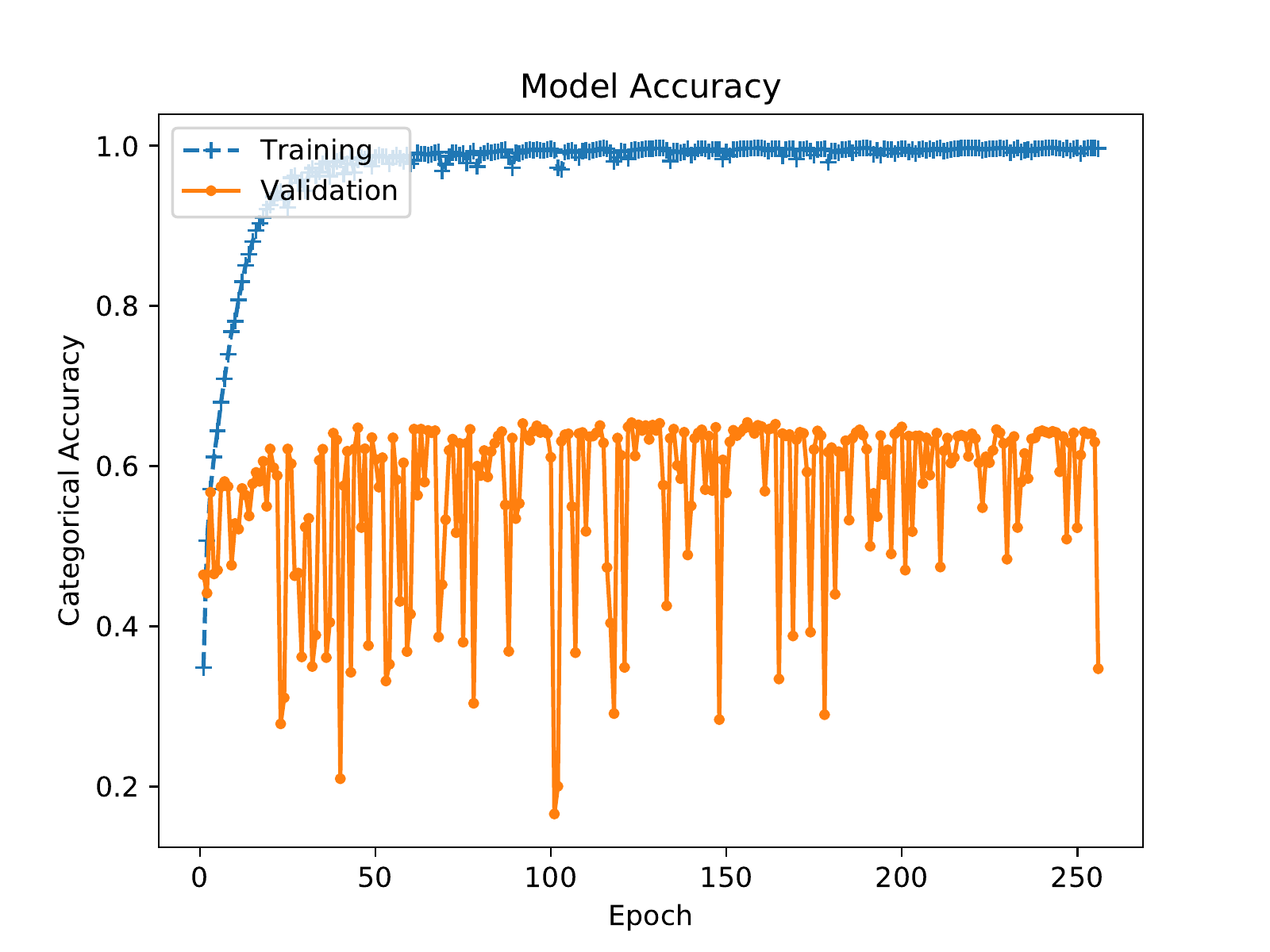}}
  \centerline{(b) DenseNet169}\medskip
\end{minipage}  
  
\caption{Training and validation accuracy versus epoch of (a) VGG16 (b) DenseNet169 DNN model. }
\label{fig:trainplt}
\end{figure}

\begin{table}[t]
  \centering
    \begin{tabular}{ | l | l | l | l |}
    \hline
    Architecture   & Parameters     & Accuracy   & Run Time (ms)  \\ \hline
    VGG16          & 16296K         & \textbf{0.671}  & 1136.473         \\
    VGG19          & 21606K         & 0.666           & 1357.670         \\
    ResNet50       & 33034K         & 0.618           & 1643.625         \\
    DenseNet121    & 9143K          & 0.493           & 1921.401         \\
    DenseNet169    & 15404K         & 0.292           & 2953.563         \\
    MobileNet      & 5335K          & 0.479           & \textbf{603.593} \\
    MobileNetV2    & 8558K          & 0.533           & 782.538   \\ \hline
    From Scratch   & N/A            & 0.648           & 684.758          \\ 
    \hline
    \end{tabular}
  \caption{Training results among different existing DNN architectures and the built-from-scratch model. Accuracy is calculated from 3589 test samples and Run Time is the time consumed to predict these samples.}
  \label{table:modelsel}
\end{table}
Furthermore, to understand the training process, we plot the training accuracy and validation accuracy of each epoch in Fig. \ref{fig:trainplt}. It can be seen that both training and validation accuracy of VGG16 quickly converge after epoch is larger than 50. Nevertheless, for a more sophisticated model such as DenseNet shown in Fig. \ref{fig:trainplt} (b), the validation accuracy does not converge despite the training accuracy rise to near 1 in a few epochs. This is an example of an over-fitted model.

\subsection{Confusion Matrix Analysis}
\label{ssec:confusion}

\begin{figure}[t]
  \centering
  \centerline{\includegraphics[width=8.5cm]{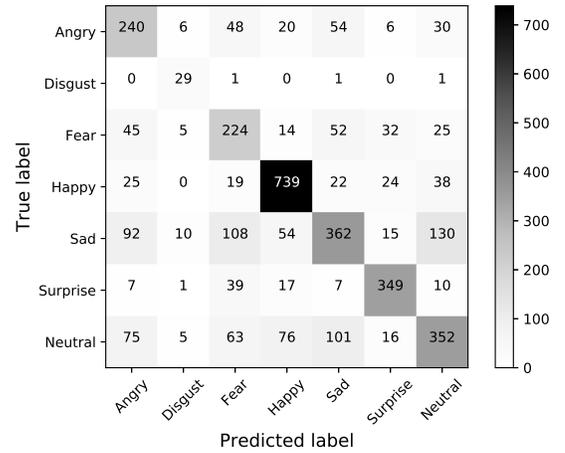}}
\caption{Confusion Matrix plotted from the testing result.}
\label{fig:confusion}
\end{figure}

To further analyze our model, we plot the confusion matrix from the prediction result of the testing data. As shown in Fig. \ref{fig:confusion}, each column of the matrix represents the instances in a predicted class while each row represents the instances in an actual class. This specific table allows us to further analyze the performance of each class: we observed that this system has the highest accuracy in prediction 'happy' emotion. However, the result is likely wrong when the system predicts a facial expression as 'sad'.\\
Given a confusion matrix $\mathbf{C}$, measurements can be defined to quantitatively analyze the performance of a model. Here, we define $Recall$ and $Precision$ as follows:
\begin{equation}
\nonumber \\ Recall_i = P(\hat{y}_t=i|y_t=i) = C_{ii}/\sum_{j} C_{ij}
\end{equation}
\begin{equation}
Precision_i = P(y_t=i|\hat{y}_t=i) = C_{ii}/\sum_{j} C_{ji},
\end{equation}
where $i$ denotes the label index. This means, precision is the accuracy where the algorithm correctly predicted class $i$ out of all classes where the algorithm predicted $i$, whereas recall  is the accuracy where the algorithm correctly predicted $i$ out of all of the cases which are labeled as $i$.\\
As demonstrated in Table. \ref{table:recallprecise}, both precision and recall are very high for 'happy'. This means that users can be confident when the system gives a 'happy' prediction. As for 'fear', the precision rate is the lowest among all labels. This number somehow implies that many emotions are mis-predicted as 'fear' (especially for 'sad' label).

\begin{table}[htb]
  \centering
    \begin{tabular}{ | l | l | l | l | l |}
    \hline
    label          & Precision      & Recall              \\ \hline
    Angry          & 0.496          & 0.594               \\
    Disgust        & 0.518          & \textbf{0.906}      \\
    Fear           & 0.446          & 0.564               \\
    Happy          & \textbf{0.803} & 0.852               \\
    Sad            & 0.604          & 0.470               \\
    Surprise       & 0.790          & 0.812               \\
    Neutral        & 0.601          & 0.512               \\
    \hline
    \end{tabular}
  \caption{Recall (True-Positive) rate and Precision rate.}
  \label{table:recallprecise}
\end{table}

\section{Conclusions}
\label{sec:foot}

To sum up, we successfully implemented a real-time image recognition system including its UI design. The main recognition algorithm is achieved by using data-driven deep learning model along with properly choosing DNN architecture and training procedure. Also, several side algorithms are deployed to increase the smoothness of this system.

Table. \ref{table:jobpartition} briefly shows the job partition of this final project: In the early stage, Ching-Da searched for available databases with proper label while Li-Heng constructed the "shell" of this system, including webcam access and result display. Later, the model training part and the fine-tuning part are mostly completed collectively.

\begin{table}[h]
  \centering

    \begin{tabular}{ | l | l | l | l | l |}
    \hline
    Item           & Li-Heng          & Ching-Da            \\ \hline
    WebCam/UI      & O                &                     \\
    Database       &                  & O                   \\
    DL-Model       & O                & O                   \\
    Pre-processing & O (scene-change) & O (de-noise)        \\
    Report         & O                & O                   \\
    \hline
    \end{tabular}
  \caption{Job partition of this project.}
  \label{table:jobpartition}
\end{table}

%


\bibliographystyle{IEEEbib}
\bibliography{strings,refs}

\begin{thebibliography}{10}

\bibitem{Li2013}
Du-Hsiu Li, Hsueh-Ming Hang, and Yu-Lun Liu,
\newblock ``Virtual view synthesis using backward depth warping algorithm,''
\newblock in {\em 2013 Picture Coding Symposium ({PCS})}. Dec. 2013, {IEEE}.

\bibitem{Liu2014}
Yu-Lun Liu and Hsueh-Ming Hang,
\newblock ``Background modeling using depth information,''
\newblock in {\em Signal and Information Processing Association Annual Summit
  and Conference ({APSIPA}), 2014 Asia-Pacific}. Dec. 2014, {IEEE}.

\bibitem{Zhang14}
Guo Y, Tian Y, Gao X, and Zhang X,
\newblock ``Micro-expression recognition based on local binary patterns from
  three orthogonal planes and nearest neighbor method,''
\newblock {\em Neural Networks (IJCNN), International Joint Conference on.
  IEEE}, pp. 3473--79, 2014.

\bibitem{AP91}
Matthew~A. Turk and Alex~P. Pentland,
\newblock ``Face recognition using eigenfaces,''
\newblock {\em Computer Vision and Pattern Recognition. Proceedings, CVPR'91.,
  IEEE Computer Society Conference on. IEEE}, pp. 586--91, 1991.

\bibitem{tu2018content}
Zhengzhong Tu, Tongyu Zong, Xueliang Xi, Li~Ai, Yize Jin, Xiaoyang Zeng, and
  Yibo Fan,
\newblock ``Content adaptive tiling method based on user access preference for
  streaming panoramic video,''
\newblock in {\em 2018 IEEE International Conference on Consumer Electronics
  (ICCE)}. IEEE, 2018, pp. 1--4.

\bibitem{opencv_library}
G.~Bradski,
\newblock ``{The OpenCV Library},''
\newblock {\em Dr. Dobb's Journal of Software Tools}, 2000.

\bibitem{jaffe97}
M.~Kamachi M.~J.~Lyons and J.~Gyoba,
\newblock ``Japanese female facial expressions (jaffe),''
\newblock {\em Database of digital images}, 1996.

\bibitem{feiface}
C.~E. Thomaz and G.~A. Giraldi,
\newblock ``{FEI} face database,''
\newblock {\em Available: https://fei.edu.br/~cet/facedatabase.html}.

\bibitem{kaggle13}
P.~L.~Carrier I.~J.~Goodfellow, D.~Erhan et~al.,
\newblock ``Challenges in representation learning: A report on three machine
  learning contests,''
\newblock {\em in International Conference on Neural Information Processing},
  pp. 117--124, 2013.

\bibitem{liu2019cyclicgen}
Yu-Lun Liu, Yi-Tung Liao, Yen-Yu Lin, and Yung-Yu Chuang,
\newblock ``Deep video frame interpolation using cyclic frame generation,''
\newblock in {\em Proc. AAAI}, 2019, pp. 8794--8802.

\bibitem{vgg}
K.~Simonyan and A.~Zisserman,
\newblock ``Very deep convolutional networks for large-scale image
  recognition,''
\newblock {\em CoRR}, vol. abs/1409.1556, 2014.

\bibitem{res}
Kaiming He, Xiangyu Zhang, Shaoqing Ren, and Jian Sun,
\newblock ``Deep residual learning for image recognition,''
\newblock in {\em arXiv prepring arXiv:1506.01497}, 2015.

\bibitem{dense}
Gao Huang et~al.,
\newblock ``Densely connected convolutional networks,''
\newblock {\em IEEE conference on computer vision and pattern recognition},
  2017.

\bibitem{mobile}
Andrew~G. Howard et~al.,
\newblock ``Mobilenets: Efficient convolutional neural networks for mobile
  vision applications,''
\newblock {\em arXiv preprint arXiv:1704.04861}, 2017.

\end{thebibliography}

\end{document}